\pdfoutput=1
\documentclass[11pt]{article}

\usepackage[final]{acl}
\usepackage{times}
\usepackage{latexsym}
\usepackage{hyperref}
\usepackage{multirow}
\usepackage{tabularx}
\usepackage{booktabs}
\usepackage{amsmath}
\usepackage{graphics}
\usepackage[T1]{fontenc}
\usepackage[utf8]{inputenc}
\usepackage{microtype}
\usepackage{inconsolata}
\usepackage{graphicx}
\usepackage{xcolor, colortbl}
\definecolor{lightyellow}{HTML}{fef9e7}

\title{GraphLSS: Integrating Lexical, Structural, and Semantic Features\\for Long Document Extractive Summarization}

\author{
  \textbf{Margarita Bugue\~no\textsuperscript{1,2}},
  \textbf{Hazem Abou Hamdan\textsuperscript{2}},
  \textbf{Gerard de Melo\textsuperscript{1,2}}
\\
  \textsuperscript{1}Hasso Plattner Institute (HPI),
  \textsuperscript{2}University of Potsdam
\\
  Potsdam, Germany \\
   \normalsize  \texttt{\{margarita.bugueno, gerard.demelo\}@hpi.de}
}

\begin{document}
\maketitle
\begin{abstract}
Heterogeneous graph neural networks have recently gained attention for long document summarization, modeling the extraction as a node classification task. Although effective, these models often require external tools or additional machine learning models to define graph components, producing highly complex and less intuitive structures.
We present GraphLSS, a heterogeneous graph construction for long document extractive summarization, incorporating Lexical, Structural, and Semantic features.
It defines two levels of information (words and sentences) and four types of edges (sentence semantic similarity, sentence occurrence order, word in sentence, and word semantic similarity) without any need for auxiliary learning models. 
Experiments on two benchmark datasets show that GraphLSS is competitive with top-performing graph-based methods, outperforming recent non-graph models. We release our code on GitHub\footnotemark{}.
\end{abstract}

\footnotetext{\url{https://github.com/AbouClaude/GraphLSS}}

\section{Introduction}
Extractive document summarization condenses documents into summaries by selecting only the most relevant sentences.  
One intuitive approach is to model cross-sentence relationships using graphs.
While prior work considered homogeneous graphs \cite{tixier-etal-2017-combining, xu-etal-2020-discourse}, recent heterogeneous graph proposals have shown high effectiveness \cite{wang-etal-2020-heterogeneous, jia-etal-2020-neural}, as they define complex relationships between multiple semantic units and capture long-distance dependencies. 
Despite their success in summarizing long documents such as scientific papers, many efforts have been made to devise more effective graph constructions.
These vary in their definitions of nodes, often requiring external tools or additional machine learning models \cite{cui-etal-2020-enhancing}, and of edges, which despite being effective, may lead to complex structures that reduce the intuitiveness of the resulting graphs \cite{zhang-etal-2022-hegel}.

This paper introduces GraphLSS, a graph construction that avoids the need for external learning models to define nodes or edges. GraphLSS utilizes \textbf{L}exical, \textbf{S}tructural, and \textbf{S}emantic features, incorporating two types of nodes (sentences and words) and four types of edges (sentence order, sentences semantic similarity, words semantic similarity, and word--sentence associations). 
We limit word nodes to nouns, verbs, and adjectives for their high semantic richness \cite{10.3233/IDA-200007, xiao-carenini-2019-extractive}. Our document graphs are processed with GAT \cite{velickovic2018graph} models on two summary benchmarks, PubMed and arXiv, which are preprocessed and labeled by us. 

Our contributions are: \textbf{i.} A novel heterogeneous graph construction using lexical, structural, and semantic features, \textbf{ii.} State-of-the-art results on both benchmarks compared to previous graph strategies and recent non-graph methods, \textbf{iii.} 
We share our code, including calculated extractive labels and graph-data creation pipeline, on GitHub\footnotemark[\value{footnote}] for reproducibility and collaboration.

\section{Previous Work}

\paragraph{Graph Structure}
Developing an effective graph structure for summarization has been challenging, leading to a proliferation of diverse approaches. \citet{wang-etal-2020-heterogeneous} proposed connecting sentence nodes to word nodes by establishing undirected associations with the contained words.
Subsequently, \citet{jia-etal-2020-neural} extended this by introducing named entity nodes and three other edge types:
directed edges for tracking subsequent named entities and words in a sentence, directed edges for entities and words within a sentence, and undirected edges for sentence pairs with trigram overlap.

Topic-GraphSum \cite{cui-etal-2020-enhancing} was one of the first attempts to apply graph strategies to long document extractive summarization. It integrated a joint neural topic model to discover latent topics in a document, defining these as intermediate nodes to capture inter-sentence relationships across various genres and lengths.
SSN \cite{cui-hu-2021-sliding} defined a sliding selector network with dynamic memory. SSN splits a given document into multiple segments, encodes them with BERT \cite{devlin-etal-2019-bert}, and selects salient sentences. Instead of representing the document as a graph, it uses a graph-based memory module, updated iteratively with a GAT \cite{velickovic2018graph}, to allow information to flow across different windows.
HeterGraphLongSum \cite{phan-etal-2022-hetergraphlongsum} utilized words, sentences, and passages as nodes, while considering undirected edges for words in sentences, and directed edges for words in passages and passage to sentences. Instead of pre-trained embeddings, it used CNNs and bidirectional LSTMs for node encoding, yielding outstanding results. MTGNN-SUM \cite{doan-etal-2022-multi} achieved similar results by capturing both inter and intra-sentence information when combining a homogeneous graph of sentence nodes with a heterogeneous graph of words and sentences, as in \citet{wang-etal-2020-heterogeneous}.

Recent studies underscore the importance of structural information in long document summarization.
HEGEL \cite{zhang-etal-2022-hegel} modeled documents as hypergraphs, capturing semantic edges like keyword coreference, section structure, and latent topics. 
CHANGES \cite{zhang-etal-2023-contrastive-hierarchical} introduced a sentence--section hierarchical graph, creating fully connected subgraphs for sentences and sections, and linking sentences to their sections. 

\paragraph{Sentence Labeling}
There is no consensus on generating extractive ground truth labels. Most previous work \cite{jia-etal-2020-neural, zhang-etal-2022-hegel, wang2024study} used the \citet{Nallapati_Zhai_Zhou_2017} greedy approach without specifying the ROUGE n-gram level, which significantly impacts sentence classifier performance. 
Some methods \cite{wang-etal-2020-heterogeneous, doan-etal-2022-multi, zhang-etal-2023-contrastive-hierarchical} selected sentences by maximizing the ROUGE-2 score against the gold summary \citet{liu-lapata-2019-text}, while others \cite{cui-etal-2020-enhancing, cui-hu-2021-sliding, phan-etal-2022-hetergraphlongsum} used pre-labeled benchmarks \cite{xiao-carenini-2019-extractive} which maximized ROUGE-1. Conversely, \citet{cho-etal-2022-toward} maximized the average of ROUGE-1 and ROUGE-2. 

\section{GraphLSS}
\label{s:method}

\paragraph{Graph Construction}
We propose a heterogeneous model that represents documents as undirected graphs, $G = (V, E)$. We use sentences and words as nodes, $V = V_\text{s} \cup V_\text{w}$, and four edge types to capture Lexical, Structural, and Semantic features, as $E = \{E_\text{ns}, E_\text{ss}, E_\text{ws}, E_\text{ww}\}$. 
Here, $V_\text{s}$ corresponds to the $n$ sentences in the document, and $V_\text{w}$ denotes the set of $m$ unique words of the document, limited to the most semantically rich ones, nouns, verbs, and adjectives.
Conversely, boolean edges $E_\text{ns}$ indicate the sentence occurrence order in documents, and $E_\text{ss}$ includes sentence pair edges, weighted by cosine similarity, within a predefined window size to account for local similarity and prevent dense graphs.
$E_\text{ws}$ denotes words in sentence edges (tf-idf weighted), and $E_{ww}$ are word-pair edges using cosine similarity.
  
\paragraph{Adaptive Class Weights}
Our graphs are processed by a heterogeneous GAT \cite{velickovic2018graph} followed by a sentence node classifier to conduct the extractive summarization. Since the extractive ground truth labels for long documents are highly imbalanced, we optimize the model using weighted cross-entropy loss. 
We assign initial class weights to relevant and irrelevant sentences, employing adaptive class weights for the relevant class and static weights for non-summary sentences:
\begin{equation}
  \label{eq:acw}
  \lambda^{i+1} = \lambda^{i} - \left(\tau -\frac{\tau}{\log(\tau)}\right),
\end{equation}
with $\tau$ the portion of sentences predicted as relevant for the summary over all the existing sentences.

\section{Experiments}

\paragraph{Datasets}
We use two publicly available benchmarks for long document summarization, PubMed and arXiv \cite{cohan-etal-2018-discourse}. Both comprise scientific English articles and are widely used by previous work. Statistics are given in \autoref{a:data}.

\paragraph{Extractive Labels}
\label{s:prepro}
Extractive labels are obtained by greedily optimizing the ROUGE-1 score, an intuitive and widely used method that allows us to label more sentences as relevant than alternative strategies. 
In the data from \citet{xiao-carenini-2019-extractive}, we identified substantial errors in the sentence tokenization. Hence, we independently preprocessed and labeled the data, removing duplicates, empty samples, and instances where abstracts exceeded source document lengths. We also replaced special characters (e.g., \texttt{\textbackslash\textbackslash}, \dots, \texttt{»}, \texttt{``''}, \texttt{\textbackslash n}) with blanks. 
We applied sentence tokenization using NLTK and merged particularly short sentences with their preceding ones (cf.~\autoref{a:data}). For word node definitions, we converted sentence text to lowercase, removing non-ASCII characters, punctuation, and stopwords. The resulting graph datasets are described in \autoref{tab:graph_statistics}.

\begin{table}[ht]
\centering
\resizebox{\linewidth}{!}{
\begin{tabular}{p{1.2cm}ccccccr}
\toprule
 & $V_\text{s}$  & $V_\text{w}$ & $E_\text{ns}$ & $E_\text{ss}$ & $E_\text{ws}$ & $E_\text{ww}$ & Disk \\
\cmidrule(lr){2-3}
\cmidrule(lr){4-7}
\cmidrule(lr){8-8}
\multirow{2}{*}{PubMed}          & 80 & 156 & 80 & 60 & 738 & 27 & 365 KB\\
\hspace{0.5cm} & 34\% & 66\%    & 9\% & 6\% & 82\% & 3\% & \\
\cmidrule{2-8}
\multirow{2}{*}{arXiv}           & 123 & 154 & 122 & 50 & 879 & 10 & 421 KB\\
\hspace{0.5cm}  & 44\% & 56\%   & 11\% & 5\% & 83\% & 1\% &\\ 
\bottomrule
\end{tabular}
}
\caption{GraphLSS statistics and average disk usage.}
\label{tab:graph_statistics}
\end{table}

\vspace{-0.3cm}
\paragraph{Comparison Methods} 

For a more detailed comparative analysis with the models that achieved the best benchmark results (Topic-GraphSum, SSN, and HeterGraphLongSum), we also executed our model using the preprocessed data and sentence-level relevance labels provided by \citet{xiao-carenini-2019-extractive}.
We also include results from recent non-graph extractive summarizers in \autoref{t:results} for reference: Lodoss \cite{cho-etal-2022-toward} learns sentence representations through simultaneous summarization and section segmentation, Topic-Hierarchical-Sum \cite{wang2024study} uses local topic information and hierarchical extraction modules, and LOCOST \cite{le-bronnec-etal-2024-locost} is an abstractive summarization model based on state-space models for conditional text generation.

\paragraph{Experimental Setup}

We trained a GAT model \cite{velickovic2018graph} with 4 attention heads and 1--2 hidden layers, minimizing binary cross-entropy loss with adaptive class weights (\autoref{eq:acw}). We initialized word nodes using \href{https://nlp.stanford.edu/projects/glove/}{GloVe Wiki-Gigaword 300-dim.\ embeddings} \cite{pennington-etal-2014-glove} and pre-trained SBERT (\texttt{All-MiniLM-L6-v2}) embeddings for sentence nodes \cite{reimers-gurevych-2019-sentence}. Further details are given in \autoref{a:exps}.

\section{Results \& Analysis}

\autoref{t:results} presents the results of different approaches, with graph-based models listed first, followed by non-graph baselines as reference, and our results. 
ROUGE-1/-2/-L F1-score is measured to assess the informativeness and fluency of the summaries. 

\begin{table*}[!ht] 
\small
\centering
\begin{tabularx}{\linewidth}{p{8cm}rrrrrr} 
\toprule
& \multicolumn{3}{c}{PubMed} & \multicolumn{3}{c}{arXiv} \\
\cmidrule(lr){2-4}
\cmidrule(lr){5-7}
Model & R-1 & R-2 & R-L & R-1 & R-2 & R-L \\
\midrule
\rowcolor{lightyellow}
Oracle \cite{xiao-carenini-2019-extractive}             & 55.05 & 27.48 & 38.66 & 53.88 & 23.05 & 34.90  \\ 
$\rightarrow$ Topic-GraphSum \cite{cui-etal-2020-enhancing} $\dagger$ & $\star$48.85 & \underline{21.76} & 35.19 & \underline{46.05} & $\star$19.97 & 33.61 \\  
$\rightarrow$  SSN \cite{cui-hu-2021-sliding} $\dagger$                & 46.73 & 21.00 & 34.10 & 45.03 & 19.03 & 32.58 \\  
$\rightarrow$  HeterGraphLongSum \cite{phan-etal-2022-hetergraphlongsum} $\dagger$ & $\star$48.86 & $\star$22.63 & $\star$44.19 & $\star$47.36 & \underline{19.11} & $\star$41.47 \\  
$\rightarrow$  MTGNN-SUM \cite{doan-etal-2022-multi}                   & 48.42 & 22.26 & 43.66 & 46.39  & 18.58 & 40.50 \\  
$\rightarrow$  HEGEL \cite{zhang-etal-2022-hegel}                      & 47.13 & 21.00 & 42.18 & 46.41 & 18.17 & 39.89 \\
$\rightarrow$  CHANGES \cite{zhang-etal-2023-contrastive-hierarchical} & 46.43 & 21.17 & 41.58 & 45.61 & 18.02 & 40.06 \\ 
\midrule
$\rightarrow$  Lodoss \cite{cho-etal-2022-toward} & 49.38 & 23.89 & 44.84 & 48.45 & 20.72 & 42.55\\
$\rightarrow$  Topic-Hierarchical-Sum \cite{wang2024study} & 46.49 &	20.52 &	42.06 & 45.84 & 19.03 & 40.36 \\
$\rightarrow$  LOCOST \cite{le-bronnec-etal-2024-locost} & 45.70 & 20.10 &	42.00 & 43.80 & 17.00 &	39.70 \\
\midrule
\rowcolor{lightyellow}
Our Oracle & 60.58 & 36.91 & 55.32 & 63.57 & 30.40 & 54.10 \\ 
$\rightarrow$  GraphLSS + Labels by \citet{xiao-carenini-2019-extractive} $\dagger$ 
& \underline{47.85} & 21.74 & \underline{42.22} & 45.91	& 18.35	& \underline{40.07} \\
$\rightarrow$  GraphLSS + Our labels & \textbf{51.42} & \textbf{24.32} & \textbf{49.48} & \textbf{55.14} & \textbf{23.00} & \textbf{50.83} \\ \bottomrule
\end{tabularx}
 \caption{ROUGE F1 results with scores from respective papers. Models using data from \citet{xiao-carenini-2019-extractive} are marked with $\dagger$ for direct comparison. 
 Best results are marked with $\star$, and second-best are underlined. Bold highlights the GraphLSS improvement, whose results are averaged over 3 runs.
}
 \label{t:results}
\end{table*}

\paragraph{Summarization Results}

GraphLSS significantly outperforms all compared approaches in ROUGE-1/-2/-L scores on PubMed and arXiv, effectively identifying relevant sentences in highly imbalanced settings (\autoref{eq:acw}). These results are based on our preprocessing and labeling. The Oracle results using our labels also greatly exceed those achieved with the data by \citet{xiao-carenini-2019-extractive}. 
With the latter labels, GraphLSS remains competitive (especially regarding ROUGE-L), despite not relying on auxiliary tools and models.
This demonstrates close alignment with reference summaries in terms of the longest common subsequence, while alternative approaches yield contaminated summaries.
Only HeterGraphLongsum surpasses GraphLSS by using CNN and LSTM networks to learn text embeddings from scratch, whereas we leverage pre-trained embeddings to reduce memorization and bias. 
These results also suggest that GraphLSS, even with pre-labeled data, outperforms recent non-graph models. 
Other graph methods are included for reference only, as they are not directly comparable due to the use of different labeling strategies in part requiring extrinsic resources.

\paragraph{Labeling Impact}
\autoref{t:results} shows that summarization results can vary significantly, depending not only on the graph construction and model but also on the strategy used for generating extractive labels. This crucial aspect has been overlooked in related work, which often focuses on ROUGE results without considering whether the corresponding methods are using the same labeling approach.
Moreover, preprocessing steps prior to label calculation can also affect the results. Although \citet{xiao-carenini-2019-extractive} and our study aimed to maximize the ROUGE-1 score, our labels differ significantly. Comparable setups are a requirement to accurately assess the advantages of models.

\paragraph{Balance of Precision \& Recall} 
\autoref{t:pre_re} shows that a two-layer heterogeneous GAT outperforms a single-layer GAT on both datasets, indicating the benefit of extended message passing across the multiple semantic units.
Additionally, previous work has not adequately addressed the balance between precision and recall, focusing solely on reporting the F1 score without analyzing the individual values and their implications.
Our results show that precision and recall are similar for the experiments on PubMed, 
reflecting a strong alignment between generated and gold summaries for both ROUGE-1 and ROUGE-2.  
In contrast, recall considerably exceeds precision on the arXiv dataset, suggesting our model retrieves relevant information but generated summaries still harbors additional text. This effect is more pronounced with a two-layer GAT. 
Interestingly, this discrepancy is not observed when using the pre-labeled data from \citet{xiao-carenini-2019-extractive}, where precision and recall are balanced, though lower. 
This suggests that the observed differences are due to data labeling artifacts rather than the graph construction or the GAT model, emphasizing our earlier discussion.

\begin{table}[!ht] 
\small
\centering
\resizebox{\linewidth}{!}{
\begin{tabular}{p{1.1cm}lrrrrrrp{0.7cm}}
\toprule
 & & \multicolumn{3}{c}{ROUGE-1} & \multicolumn{3}{c}{ROUGE-2} & \multirow{2}{\hsize}{Time $\left[\text{h}\right]$}\\ 
\cmidrule(lr){3-5}
\cmidrule(lr){6-8}
Dataset & $L$ & P & R & F1 & P & R & F1 & \\
\midrule
\multirow{3}{*}{PubMed} & 1 & 49.75 & 50.00 & 49.92 & 22.61 & 24.71 & 23.17         & 19.9\\  
                        & 2 & 52.59	& 50.11 & 51.42 & 23.91	& 23.82	& 24.32         & 26.1\\ 
                        & 2 $\dagger$ & 46.43	& 49.42	& 47.85 & 22.42 & 21.14 & 21.74 & 26.2\\ 
\midrule
\multirow{3}{*}{arXiv}  & 1 & 45.66	& 66.68 & 54.23 & 17.14 & 30.20 & 22.31         & 22.8\\
                        & 2 & 45.20 & 71.04 & 55.14 & 17.02 & 35.74	& 23.00         & 31.9\\
                        & 2 $\dagger$  & 44.88 & 47.04 & 45.91 & 19.96 & 16.99 & 18.35    & 32.2 \\ 
\midrule
\end{tabular}
}
 \caption{ROUGE scores as precision (P) and recall (R). $L$ indicates the number of GAT layers, and $\dagger$ marks results using data from \citet{xiao-carenini-2019-extractive}. } 
 \label{t:pre_re}
\end{table}

\vspace{-0.4cm}
\paragraph{Resources} 
While arXiv articles are around 50\% longer than those in PubMed, the graph size increases only by 15\% in nodes and 75\% in edges. 
Since nodes are represented by high-dimensional vectors and edges by single values, GraphLSS disk usage is mainly determined by node count, resulting in a 15\% increase for arXiv (56 KB per graph).
Such an increase is also reflected in the GAT training time (\autoref{t:pre_re}). Conversely, increasing model complexity from 1 to 2 GAT layers raises training time by 32\% on PubMed and 40\% on arXiv. 

\paragraph{Ablation Study}
We conducted an ablation study on PubMed to assess the contributions of each edge type (\autoref{t:ablation}). The results indicate that word-in-sentence edges have the highest impact on GraphLSS performance, as their removal significantly reduces ROUGE scores. This highlights the importance of cross-granularity interactions for effective document representation. 
Notably, around 80\% of node associations are discarded when removing such edges, isolating words and sentences into separate components.
Sentence edges are also important, with a comparable effect on ROUGE. However, sentence similarity edges are relatively more influential than sentence order ones due to their lower edge count.
In turn, word similarity edges have the least impact, reflecting their low representation in the graph (only 3\%; \autoref{tab:graph_statistics}).

\begin{table}[!ht] 
\small
\centering
\resizebox{0.95\linewidth}{!}{
    \begin{tabular}{lrrr}
    \toprule
     & R-1 & R-2 & R-L \\ 
    \midrule
    \rowcolor{lightyellow}
    GraphLSS                            & 51.42 & 24.32 & 49.48 \\
    (--) Word in Sentence $E_{ws}$      & 47.91 & 21.96 & 46.02 \\
    (--) Sentence Similarity $E_{ss}$   & 48.87 & 22.39 & 46.68 \\
    (--) Sentence Occurrence $E_{ns}$   & 48.99 & 22.41 & 46.65 \\   
    (--) Word Similarity $E_{ww}$       & 50.84 & 23.78 & 48.80 \\
    \bottomrule
    \end{tabular}
    }
\caption{Ablation study on PubMed. Results were obtained by removing one specific edge type.
} 
\label{t:ablation}
\end{table}

\vspace{-0.3cm}
\section{Conclusions}

We introduced GraphLSS, a heterogeneous graph for long document summarization incorporating lexical, structural, and semantic features.
Experiments on PubMed and arXiv highlight the impact of extractive labels due to their inherent imbalance. 
GraphLSS proves competitive with top-performing graph-based methods and outperforms recent non-graph models by using a greedy labeling strategy and adaptive weights during training. 
Future work will explore integrating an abstractive summarizer.

\section*{Limitations}

While we showed the impact and potential of GraphLSS for long document extractive summarization, there are some points to keep in mind.

Storing document graphs as a data structure obtained from the original documents (texts) involves significant additional disk usage. Previous strategies create such structures on the fly while training the underlying GNN models, and others opt for storing such graphs on disk to speed up model training. We follow the latter strategy. Therefore, the training time reported does not consider the creation of the underlying graphs.

Furthermore, our proposal was only validated on English datasets. Applying GraphLSS to other languages may yield significantly different results, since pre-trained word and sentence embeddings are required for node initialization and thus, training the heterogeneous GAT model. Analyzing this aspect would be particularly interesting for
low-resource languages.
Additionally, our experiments focus on scientific papers. Although they cover multiple scientific domains, exploring other kinds of long document, e.g., narrative and legal documents, is encouraged. Also, additional data collections should be analyzed in order to generalize our findings to broader domains.

\section*{Ethics Statement}

While extractive summaries are less prone to hallucinated content, in some instances, they may be misleading due to missing context.  
Another concern is that of possible bias during the content selection. Depending on the graph construction applied, a GAT model may favor certain types of content over others, such as popular sentences and entities with high degrees, as they might receive more attention. Thus, special care must be taken when relying on summaries to make high-stakes decisions, for example in the legal or medical domains.

Summarizing articles often involves extracting information related to trending topics, institutions, people, and other entities. Balancing the delivery of valuable summaries while respecting the privacy of these entities is essential. One strategy to alleviate such concern is anonymization, which ensures that the summary content does not reveal sensitive features. In our study, we conduct all experiments on publicly available scientific articles, and hence have forgone such anonymization.

\bibliography{anthology,custom}

\appendix

\section{Dataset Statistics}
\label{a:data}

We use two publicly available benchmarks for long document summarization, PubMed and arXiv \cite{cohan-etal-2018-discourse}. PubMed comprises biomedical scientific papers collected from \href{https://pubmed.ncbi.nlm.nih.gov/}{pubmed.ncbi.nlm.nih.gov}, while arXiv covers various scientific domain articles collected from \href{https://arxiv.org/}{arXiv.org}. The statistics of both datasets are presented in \autoref{t:data}. 

\begin{table}[!ht] \small
\newcolumntype{C}{>{\centering\arraybackslash}X}
\newcolumntype{L}{>{\raggedright\arraybackslash}X}
\newcolumntype{R}{>{\raggedleft\arraybackslash}X}
\centering
\begin{tabularx}{\linewidth}{p{4cm}rr}
\toprule
            & PubMed & arXiv \\
\midrule
\#Training  & 115,776 & 197,650 \\
\#Validation & 6,584 & 6,435 \\
\#Testing   & 6,620 & 6,439 \\
\midrule
Avg.\ \# Tokens in doc.       & 2,768 & 3,913\\
Avg.\ \# Tokens in summary    & 205   & 203\\
Avg.\ \# Sentences in doc.    & 89  & 133\\
Avg.\ \# Sentences in summary & 8   & 7\\
\midrule
\end{tabularx}
 \caption{Datasets statistics.}
 \label{t:data}
\end{table}

\subsection{Preprocessing Details}
\label{a:cleaning}

As described in \autoref{s:prepro}, we removed duplicate and empty documents and instances where the article is shorter than the corresponding summarization. Subsequently, we split the documents via NLTK's sentence tokenizer. However, since the sentence tokenizer splits text based on punctuation, this can often result in non-sensical sentences. For example, the sentence \textit{``Neptune masses can be excluded by our limits determinations (fig.1)"} results in a head sentence $S_{h}=$\textit{``Neptune masses can be excluded by our limits determinations (fig."} and a tail sentence $S_{t}=$\textit{``1)."}. In such cases, we merged tail sentences with the preceding ones to maintain text coherence.

\section{Further Experimental Details}
\label{a:exps}

\paragraph{Experimental Setup}
We trained a GAT model \cite{velickovic2018graph} with 4 attention heads, varying the number of hidden layers between 1 and 2. 
We applied Dropout after every GAT layer with a retention probability of 0.7. The final representation is fed into a sigmoid classifier.
We initialized word nodes using \href{https://nlp.stanford.edu/projects/glove/}{GloVe Wiki-Gigaword 300-dim.\ embeddings} \cite{pennington-etal-2014-glove} and pre-trained SBERT (\texttt{All-MiniLM-L6-v2}) embeddings for sentence nodes \cite{reimers-gurevych-2019-sentence}. 
Notably, our word nodes are restricted to the top 50,000 most frequent words in the respective dataset's vocabulary.

All experiments used a batch size of 64 samples and were trained for a maximum of 20 epochs using Adam optimization with an initial learning rate of $10^{-3}$. The training was stopped if the validation loss did not improve for 7 consecutive iterations. The objective function of each model was to minimize the binary cross-entropy loss using adaptive class weights, as described in \autoref{eq:acw}.
All experiments are based on PyTorch Geometric and conducted on an NVIDIA GeForce RTX 3050.
We share our code and labeled datasets, including the graph creation pipeline on \url{https://github.com/AbouClaude/GraphLSS}.

\begin{figure}[h]
    \centering
    \includegraphics[width=0.9\columnwidth]{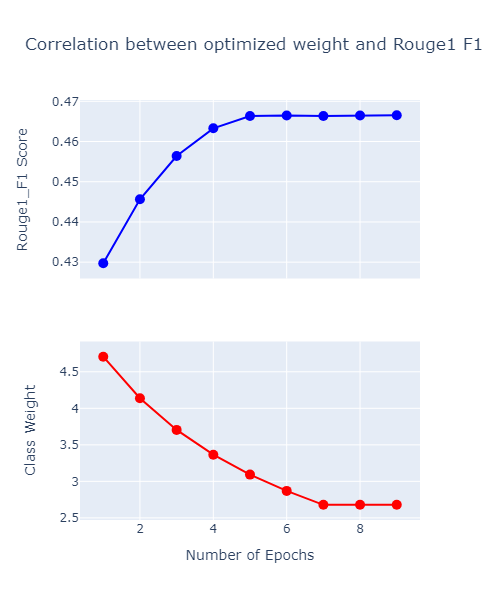} 
    \caption{Effect of adaptive class weights on PubMed. }
    \label{fig:acw}
\end{figure}

\paragraph{Adaptive Class Weights}
\autoref{fig:acw} illustrates how the adaptive class weights evolve across epochs during training. Specifically, we update the weights solely for the relevant class (summary sentences), maintaining static weights for the irrelevant class.

\section{Libraries Used} 
The experiments were conducted using the following libraries:

\begin{table}[!ht] \small
\centering
\begin{tabularx}{0.75\linewidth}{lr}
\toprule
Library & Version \\
\midrule
\texttt{nltk} &\texttt{3.8.1} \\
\texttt{pytorch} &\texttt{2.2.1} \\
\texttt{transformers} &\texttt{4.38.2} \\
\texttt{rouge} & \texttt{1.0.1}\\
\texttt{scikit-learn} &\texttt{1.3.0}\\
\texttt{torchmetrics}& \texttt{1.2.1}\\
\texttt{torch\_geometric} & \texttt{2.5.0}\\
\midrule
\end{tabularx}
 \caption{Libraries and versions.}
 \label{t:lib}
\end{table}

\end{document}